\documentclass[letterpaper, 10 pt, conference]{ieeeconf}  

\IEEEoverridecommandlockouts                              

\usepackage{amssymb}
\usepackage{float}
\usepackage{makeidx}
\usepackage{amsmath}
\usepackage{bbm}
\usepackage{epsfig}
\usepackage{epsf}
\usepackage{psfrag}
\usepackage{verbatim}
\usepackage{color}
\usepackage{multirow}
\usepackage{tabularx}
\usepackage{booktabs}
\usepackage[tight,footnotesize]{subfigure}
\usepackage{array}
\usepackage{soul}
\usepackage{footnote}
\usepackage{dblfloatfix}
\usepackage{tcolorbox}
\usepackage{caption}
\usepackage{wrapfig}
\usepackage{lipsum}

\usepackage{color, colortbl}
\usepackage[colorlinks,bookmarksnumbered,citecolor=orange,urlcolor=orange]{hyperref}
\usepackage{graphicx}
\graphicspath{{Figures}}
\DeclareGraphicsExtensions{.pdf,.png}
\usepackage{bigstrut}
\usepackage[english]{babel}

\usepackage{csquotes}

\usepackage[T1]{fontenc}
\usepackage{algorithm, setspace}
\usepackage{algpseudocode}
\usepackage{url}
\usepackage{multirow}
\usepackage{float}
\usepackage{xcolor}
\usepackage{hyperref}
 \hypersetup{
     colorlinks=true,
     linkcolor=blue,
     filecolor=blue,
     citecolor=blue,      
     urlcolor=blue,
     }

\usepackage{balance}
\usepackage{amssymb}
\let\labelindent\relax
\usepackage{enumitem}
\usepackage{cuted}

\usepackage[
    backend=biber,
    hyperref=true,
    url=false,
    isbn=false,
    doi=false,
    backref=false,
    style=ieee,
    natbib=true,
    citestyle=numeric-comp,
    sorting=none,
    block=none,
    maxbibnames=1,
    minbibnames=1
]{biblatex}
\renewcommand{\bibfont}{\small}
\title{\LARGE \bf

Unsupervised Discovery of Failure Taxonomies from Deployment Logs
}

\author{Aryaman Gupta$^{1\dagger}$, Yusuf Umut Ciftci$^{1,2\dagger}$, Somil Bansal$^1$ %
\thanks{$\dagger$ Equal contribution. $^{1}$Department of Aeronautics and Astronautics, Stanford University, USA. {\tt\footnotesize \{aryamann, somil\}@stanford.edu.}}%
\thanks{$^{2}$Electrical Engineering Department, USC, USA. {\tt\footnotesize yciftci@usc.edu.}}%
\thanks{We gratefully acknowledge research support from the DARPA ANSR program, NSF CAREER program (2240163), Precourt Institute for Energy, and the Sustainable Mobility Center at Stanford University.}
}

\begin{document}

\maketitle
\thispagestyle{empty}
\pagestyle{empty}

\begin{strip}
\begin{minipage}{\textwidth}\centering
\vspace{-6.5em}
\includegraphics[width=\textwidth]{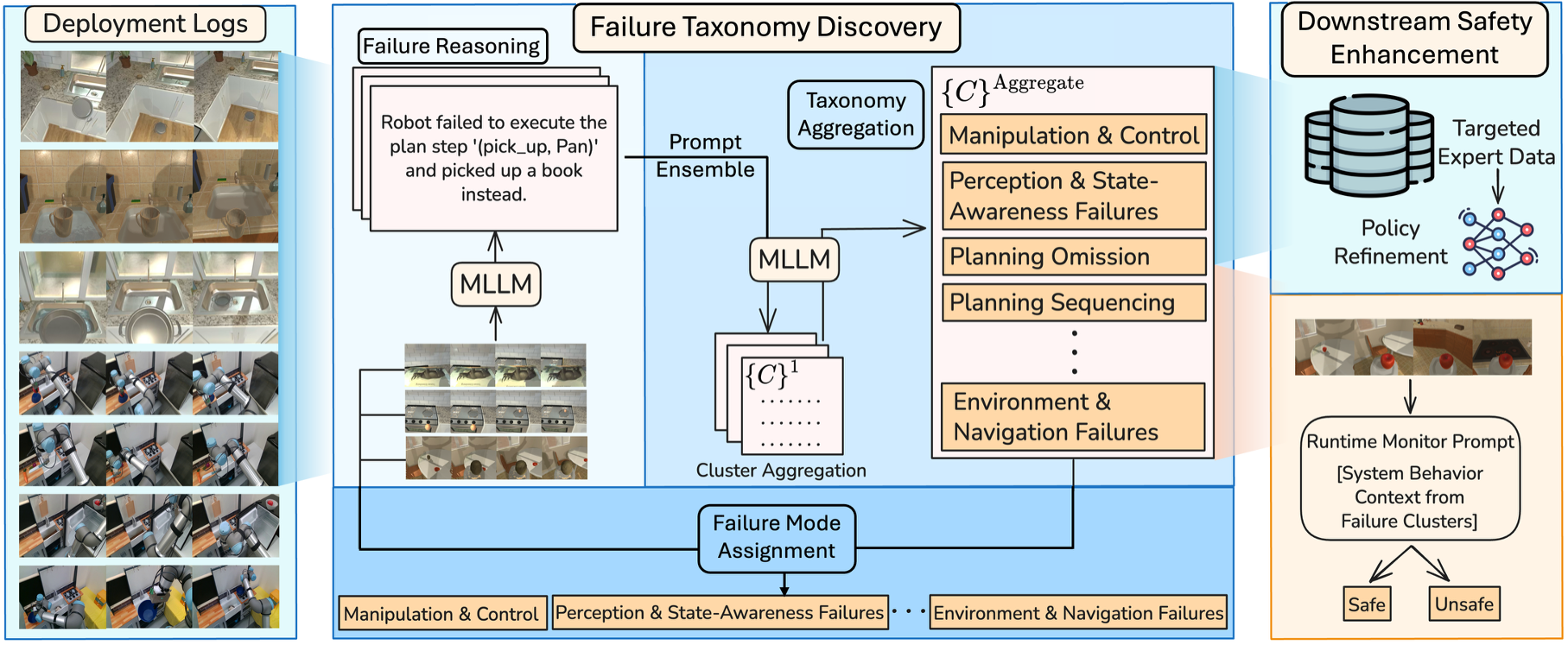}
\captionof{figure}{
A framework for unsupervised failure taxonomy discovery from deployment logs.
Given failure-centered multimodal trajectories, our method first infers structured failure explanations and then aggregates them in a semantic reasoning space to identify recurring failure modes, forming a taxonomy. This resulting failure taxonomy enables closed-loop safety improvements, including targeted data collection and context-aware runtime failure monitoring.}
\label{fig:front figure}
\vspace{-1em}
\end{minipage}
\end{strip}

\begin{abstract}

As robotic systems become increasingly integrated into real-world environments, ranging from autonomous vehicles to household assistants, they inevitably encounter diverse and unstructured scenarios that lead to failures. While such failures pose safety and reliability challenges, they also provide rich perceptual data for improving system robustness.
However, manually analyzing large-scale failure datasets is impractical and does not scale.
In this work, we introduce the problem of unsupervised discovery of failure taxonomies from large volumes of raw failure logs, aiming to obtain semantically coherent and actionable failure modes directly from perceptual trajectories. Our approach first infers structured failure explanations from multimodal inputs using vision-language reasoning, then clusters them in the resulting semantic reasoning space, discovering recurring failure modes rather than isolated episode-level descriptions.
We evaluate our method across robotic manipulation, indoor navigation, and autonomous driving domains, demonstrating that the discovered taxonomies are consistent, interpretable, and useful in practice.
In particular, we show that structured failure taxonomies guide targeted data collection for offline policy refinement and enhance runtime failure monitoring systems. Website: {\href{https://mllm-failure-clustering.github.io/}{https://mllm-failure-clustering.github.io/}}

\end{abstract}

\section{INTRODUCTION}
\label{sec:intro}

Autonomous systems, ranging from self-driving vehicles

\noindent to household robots, are increasingly deployed in open, dynamic environments. In such unstructured settings, they are prone to failures due to unexpected interactions and long-tail edge cases. Traditional validation pipelines, often grounded in simulation or controlled testing, struggle to capture the full complexity of real-world deployment, leaving many failure modes undetected until deployment.

A promising direction for improving system robustness is to systematically learn from failures that occur during deployment. Robots naturally collect large volumes of perceptual data, including traces of failed interactions. These trajectories provide valuable insights into the underlying conditions and mechanisms that led to safety violations, brittleness, or policy errors.
Such failure analysis is often manual, with human experts sifting through these logs, as in autonomous driving \cite{xu2025wod}.
However, manually curating and analyzing large-scale failure logs is time-consuming and inherently unscalable.

In this work, we introduce the problem of \textit{unsupervised failure taxonomy discovery} from multimodal, failure-centered trajectories. Rather than analyzing failures in isolation~\cite{reflect, aha, lu2025robofac, qi2026selfrefining, pacaud2025guardian}, our goal is to automatically obtain a semantically coherent and actionable failure taxonomy directly from raw deployment logs, without predefined labels.
To address this challenge, we propose an inference procedure that first extracts structured failure explanations from perceptual trajectories using vision-language reasoning, then clusters them in the resulting semantic reasoning space. By operating on inferred failure descriptions rather than raw perceptual similarity, our method identifies recurring failure patterns across deployment logs and organizes them into interpretable categories described in natural language. Importantly, the entire process is fully unsupervised, eliminating the need for costly human annotation while preserving semantically meaningful structure.

The discovered failure taxonomies provide significant downstream benefits for improving system safety. We demonstrate their value through two key applications. First, they guide targeted data collection, enabling developers to efficiently focus training efforts on critical or underrepresented failure scenarios while reducing overall data collection costs. Second, we show that these failure modes enhance online failure monitoring systems by providing richer contextual understanding of system behavior, serving as an early-warning signal for potential system safety violations at runtime.

\textbf{Contributions.}
(1) We introduce the problem of unsupervised failure taxonomy discovery from multimodal, failure-centered trajectories.
(2) We propose a framework that extracts structured failure explanations and clusters them into semantic failure modes.
(3) We demonstrate that the discovered taxonomies provide measurable closed-loop safety benefits, improving targeted data collection and runtime failure monitoring across multiple robotic domains.

\vspace{0.5em}

\section{Related Works}
\label{sec:related}

\noindent \textbf{Semantic Clustering of Images.}
Recent work has shown vision-language models (VLMs) to be effective for semantic grouping of images. Prior methods cluster images using human-specified language criteria \cite{ictc} or automatically discover clustering criteria from image datasets \cite{clusteringbylanguage}, reducing dependence on manual annotations.
Other approaches highlight semantic differences between image sets \cite{visdiff}, or identify important subpopulations and semantically diverse subsets for training \cite{subpopulationdiscovery, dataenrichment}. 
While these methods focus on static image collections, our work addresses failure-centered, multimodal trajectories in autonomous systems. Here, the objective is not merely to group perceptually similar instances but to identify recurring failure modes from temporal deployment logs.

\vspace{0.2em}
\noindent \textbf{Text Clustering and Topic Modeling.}
Topic modeling methods fall into three broad groups. First, extensions of classical probabilistic models such as Latent Dirichlet Allocation (LDA) incorporate word embeddings to improve semantic representation \cite{topic11, topic12, topic13, topic14, topic21, topic22, topic23, topic24}.
Second, fully embedding-based approaches leverage contextualized representations from pre-trained language models \cite{topic31, topic32, topic33, topic34}. Third, hybrid methods separate cluster formation from topic representation, allowing flexible topic extraction \cite{topic41, topic42, bertopic}.
Our work differs from traditional text clustering in two key ways. First, the underlying data is multimodal trajectories rather than standalone text documents. Second, we form a taxonomy of semantic failure modes grounded in deployment logs, rather than discover abstract topics in text corpora.

\vspace{0.2em}
\noindent \textbf{Failure Mining in Autonomous Systems.}
Falsification has emerged as a prominent methodology for uncovering failures in autonomous systems. Various approaches \cite{fals1,fals2,fals3,fals4,fals5,fals6,fals7} test systems in simulated environments under conditions designed to provoke failures.
These methods are effective for identifying environmental parameters responsible for system breakdowns under controlled settings.
However, they typically operate by tuning predefined parameterizations.
Moreover, identifying the underlying causes of the resulting failure cases requires manual inspection.
In contrast, our work automates the entire pipeline, leveraging large-scale deployment datasets for unsupervised failure reasoning and categorization to form interpretable, actionable taxonomies without any human effort.

\vspace{0.2em}
\noindent \textbf{Language-based Failure Reasoning in Robotics.}
Recent works integrate LLMs into robotic systems to generate natural-language failure explanations, demonstrating that informative failure descriptions enhance diagnostic capabilities in manipulation tasks \cite{reflect, aha, lu2025robofac, qi2026selfrefining, pacaud2025guardian}.
These approaches primarily operate at the individual episode level, explaining or correcting specific failures. In contrast, we address the orthogonal problem of finding corpus-level failure structure: automatically discovering and organizing failure modes across deployment logs, without supervision.

\section{Problem Formulation}
\vspace{-0.25em}
\label{sec:task}

We formalize the problem of unsupervised discovery of failure taxonomy from multimodal perceptual recordings collected during robotic deployment.
Consider a dataset of $N$ failure-centered sequences:
\[
D = \{o_{1:K_n}^n\}_{n=1}^{N},
\]
where $o_{1:K_n}^n = (o_1^n, \ldots, o_{K_n}^n)$ denotes the observation trajectory for the $n$-th failure instance.
Each sequence contains frames preceding and following the failure time index $k_f^n \in \{1,\dots,K_n\}$.
The observation $o_{k_f^n}^n$ corresponds to the failure event, while neighboring frames provide pre-failure context and post-event consequences.

Our objective is to discover a set of $L$ semantically coherent and actionable failure modes capturing the recurring failure patterns in the deployment logs. Formally, we seek a mapping
\begin{equation*}
H : D \mapsto \left\{ C_l = \left( s_{l},\, d_{l},\, \kappa_{l},\, D_{l},\, f_{l} \right) \right\}_{l=1}^{L},
\end{equation*}
where each cluster $C_l$ represents a failure mode and is characterized by:
(i) a natural language name $s_l$,
(ii) a short textual description $d_l$,
(iii) representative keywords $\kappa_l$,
(iv) a subset of sequences $D_l \subset D$ assigned to that failure mode, and
(v) a frequency $f_l = |D_l|/|D|$.

Crucially, the failure modes are not predefined and must be obtained from raw deployment data. The goal is not merely to group perceptually similar sequences, but to recover semantically coherent failure modes that explain why failures occur across episodes.
For instance, a cluster might be $C_l$ = \texttt{Rear-End Collisions: Insufficient Following Distance}, where every trajectory in $D_l$ involves an autonomous vehicle failing to maintain safe headway from a leading vehicle.

\section{Method}
\label{sec:method}

Inferring the cause of failure in a robot trajectory is a complex task that requires understanding the robot's environment, the agent's actions, interactions with other agents, and their consequences.
Performing such analysis at scale demands automated systems capable of extracting structured semantic signals from raw perceptual data and reasoning over recurring patterns.
Our approach proceeds in three stages: (1) \textbf{failure reasoning} from perceptual sequences, (2) \textbf{failure taxonomy discovery} by aggregating the explanations into semantically coherent failure modes, and (3) \textbf{assignment} of each trajectory to the failure modes, as demonstrated in Fig.~\ref{fig:front figure}.

\subsubsection{\textbf{Semantic Observation Downsampling}}
To compactly encode each failure trajectory while preserving causal context, we perform frame-level similarity-based downsampling centered on the failure event.
For each failure sequence \( o_{1:K}^n \), let the failure occur at index \( k_f \). We consider a temporal window of length \( T \) around the failure: \( o_{k_f - T_p : k_f + T_s}^n \), where \( T_p \) and \( T_s \) denote the number of pre- and post-failure frames, respectively, with \( T = T_p + T_s + 1 \). This captures both the pre-failure buildup and the immediate aftermath.

Let \( \phi(\cdot) \in \mathbb{R}^d \) denote the CLIP embedding function and define cosine similarity
\vspace{-0.5em}
\[
\mathrm{sim}(i,j) 
= 
\frac{\phi(o_i^n)^\top \phi(o_j^n)}
{\|\phi(o_i^n)\|\|\phi(o_j^n)\|}.
\]
\vspace{-0.5em}

We construct a compressed subsequence \( \tilde{o}_{1:M}^n \subset o_{k_f - T_p : k_f + T_s}^n \) by anchoring at the failure frame. Initialize \( i_0 = k_f \), include \( o_{i_0}^n \) and collect further frames following:
\vspace{2pt}

\textbf{Backward selection}:
\[
i_{m+1}^{(b)} 
= 
\max \left\{ j < i_m^{(b)} \;|\; \mathrm{sim}(j, i_m^{(b)}) < \tau \right\}.
\]

\textbf{Forward selection}:
\[
i_{m+1}^{(f)} 
= 
\min \left\{ j > i_m^{(f)} \;|\; \mathrm{sim}(j, i_m^{(f)}) < \tau \right\}.
\]

The process in each direction terminates when no valid index remains within the window. The selected indices are merged and ordered chronologically to form \( \tilde{o}_{1:M}^n \), \( M \le T \).
This procedure performs a bidirectional, embedding-space change-point selection centered at the failure. Frames are retained only if semantically distinct from the most recently selected frame in either direction. Consequently, temporally redundant observations are removed while preserving both the critical transitions leading to failure and the immediate post-failure context, ensuring efficient use of the VLM input context window.

\vspace{2pt}
\subsubsection{\textbf{Failure Reasoning}}
\label{sec:method_step1}
Each downsampled sequence is fed to a VLM along with a structured prompt asking it to summarize the scene and agent behavior over the trajectory and to infer a plausible failure cause \( r^n \) grounded in the observed evidence.
We adopt a Chain-of-Thought strategy \cite{cot} to encourage explicit intermediate reasoning and obtain the set of failure reasons \( \mathcal R = \{r^n\}_{n=1}^N \).

\vspace{2pt}
\subsubsection{\textbf{Failure Taxonomy Discovery via Semantic Aggregation}}
\label{sec:method_step2}
Discovering the taxonomy involves clustering failure explanations according to three primary objectives: (i) intra-cluster semantic coherence, (ii) minimal inter-cluster conceptual overlap, and (iii) comprehensive coverage of the explanation set $\mathcal{R}$. We leverage LLMs as optimizers \cite{llmasoptimizer} to perform clustering. Given a set of $N$ explanations, $\mathcal{R} = \{r^n\}_{n=1}^N$, the model generates $L$ distinct clusters $\{C_l\}$, each represented by the tuple $(s_l, d_l, \kappa_l)$ defined in the problem formulation. The number of clusters $L$ is implicitly optimized to best satisfy the stated criteria.

\vspace{2pt}
\textbf{Taxonomy Aggregation.}
Clustering $N$ failure explanations simultaneously is sensitive to prompt phrasing~\cite{sensitive_prompt}, and processing all explanations in a single pass is impractical for large $N$. To improve robustness, we adopt an ensemble-and-refine strategy. Starting from an initial clustering prompt, the LLM generates diverse rephrasings, each of which independently clusters $\mathcal{R}$ into a candidate taxonomy. We then prompt the LLM to reconcile these candidates into a single consolidated taxonomy $\{C\}^{\mathrm{Aggregate}}$, resolving inconsistencies in cluster boundaries, merging overlapping categories, and unifying semantic labels. The procedure can be viewed as test-time self-refinement~\cite{sefrefinellm, zhuang2026test, llmasoptimizer}, where multiple candidate solutions are generated in parallel and synthesized into an improved output. By reconciling diverse partitions of the explanation set, the resulting taxonomy is more comprehensive and internally consistent than any single candidate.

\vspace{2pt}
\subsubsection{\textbf{Assigning Trajectories to Failure Modes}}
\label{sec:method_step3}
Each trajectory is assigned to the discovered failure modes.
Given the obtained taxonomy $\{C\}^{\mathrm{Aggregate}}$ and an explanation $r^n$, we prompt an LLM with cluster names $s_l$, descriptions $d_l$, and keywords $\kappa_l$, and ask it to map $r^n$ to the most appropriate clusters.
Trajectories that do not align with any existing cluster are flagged as outliers; such instances may correspond to rare, ambiguous, or previously unseen failures and can guide future taxonomy refinement.

\section{Experiments}
\label{sec:experiments}

We evaluate our framework across three robotic domains: (i) long-horizon robot manipulation in kitchen tasks, (ii) real-world dashcam videos of car crashes, and (iii) vision-based autonomous navigation in indoor office environments. In each domain, our objective is to obtain an interpretable taxonomy of recurring failure modes from raw deployment logs and assess its utility for downstream tasks such as runtime monitoring and targeted data collection, when applicable.

We validate the framework's components at increasing levels of supervision. In the manipulation domain, where expert-defined taxonomies are available, we quantitatively evaluate failure-explanation accuracy, taxonomy alignment, and assignment performance. In the driving and navigation domains, where large-scale expert annotations are unavailable, we assess qualitative coherence and demonstrate practical utility through safety-critical downstream tasks.

We use Gemini 2.5 Pro for failure explanation and OpenAI o4-mini for taxonomy discovery, trajectory-to-cluster assignment, runtime monitoring, and LLM-based evaluation.
Model selection was based on empirical reasoning performance across candidate open and closed-source models.

\subsection{\textbf{Case Study 1: Robot Manipulation}}
\label{sec:manip_disc}
We first evaluate our framework in a controlled setting using RoboFail \cite{reflect}, a dataset of manipulation failures in long-horizon kitchen tasks. RoboFail provides an expert-defined taxonomy of eight distinct failure modes across 100 simulated videos spanning 10 tasks, along with 30 real-world UR5 robot demonstrations across 7 tasks. Each video includes task instructions, success conditions, failure timestamps, executed action plans, and expert-annotated failure reasons.
This enables quantitative validation of each stage of our framework against a ground-truth taxonomy before scaling to less structured real-world domains.

\paragraph{Validation of Failure Reasoning}
To infer failure reasons, we provide the VLM with perception inputs, task context, and the robot’s executed action plan. Action-level information lets the model distinguish planning-level errors from execution-level failures grounded in the same perceptual evidence.

\begin{figure}[htbp]
  \centering
  {\includegraphics[width=\columnwidth]{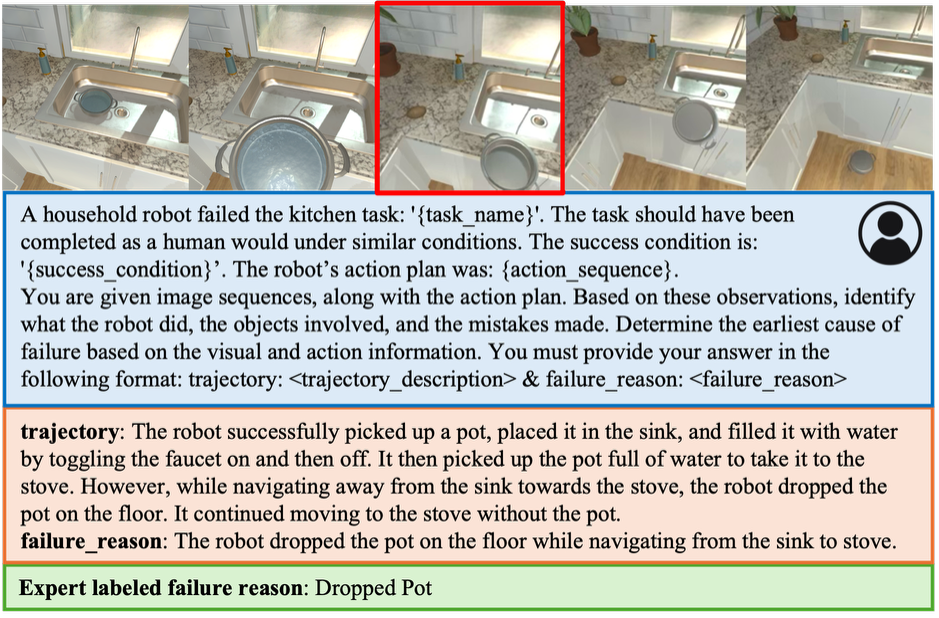}}
  \caption{A failure explanation example where the robot dropped a pot. The red-bordered frame shows failure event, blue box shows prompt, orange box shows LLM's response, and green box shows expert-labeled failure reason.}
  \label{fig:failure_inference}
  \vspace{-0.5em}
\end{figure}

Fig.~\ref{fig:failure_inference} shows a representative explanation, where the model correctly attributes failure to dropping a pot while carrying it, consistent with expert annotation.
We quantitatively evaluate agreement between inferred explanations and expert-labeled failure reasons on both simulation and real-world datasets using three complementary metrics:
(a) \textbf{Cosine Similarity (CS)} between text embeddings, 
(b) \textbf{ROUGE$_L$}, measuring lexical longest common subsequence overlap, and 
(c) \textbf{LLM-Judge (LLM-J)}, producing binary semantic agreement judgements.
All LLM-based evaluations use fixed prompts and deterministic decoding to ensure consistent scoring.
As shown in Table~\ref{tab:manip_step1}, our method achieves strong semantic alignment with expert annotations, attaining a cosine similarity of 0.60 and an LLM-J score of 0.76, demonstrating reliable failure explanation.

\begin{table}[htbp]
\vspace{0.5em}
    \centering
    \caption{Performance comparison of different VLMs on the failure reasoning task proposes Gemini 2.5 Pro is the best choice overall. The best method is highlighted in \textbf{bold}.}
    \setlength{\tabcolsep}{6pt}
    \renewcommand{\arraystretch}{1.2}
    \resizebox{\columnwidth}{!}{%
        \begin{tabular}{c|c|c|c|c|c|c}
            \toprule
            \multirow{2}{*}{\textbf{Model}}           
            & \multicolumn{3}{c|}{\textbf{Simulation}} 
            & \multicolumn{3}{c}{\textbf{Real-World}}                                                                           \\
                                                       & \textbf{CS} $\uparrow$                             
                                                       & \textbf{ROUGE$_{L}$} $\uparrow$                       
                                                       & \textbf{LLM-J} $\uparrow$ 
                                                       & \textbf{CS} $\uparrow$    
                                                       & \textbf{ROUGE$_{L}$} $\uparrow$ 
                                                       & \textbf{LLM-J} $\uparrow$ \\
            \midrule
           LLaVA-NeXT                                
            & 0.4846 & 0.2017 & 0.10 
            & 0.6305 & 0.2657 & 0.103 \\
            
            Qwen2.5-VL-7B                     
            & 0.5200 & 0.2273 & 0.26 
            & 0.6890 & 0.3570 & 0.333 \\
            
            OpenAI o4-mini                             
            & 0.5557 & 0.2422 & 0.46 
            & 0.6770 & \textbf{0.3640} & 0.370 \\
            
            Cosmos-Reason1-7B                        
            & 0.5324 & 0.1541 & 0.20 
            & \textbf{0.6930} & 0.2290 & 0.133 \\
            
            \midrule
            \textbf{Gemini 2.5 Pro} 
            & \textbf{0.6003} & \textbf{0.2589} & \textbf{0.76}    
            & 0.6280 & 0.3420 & \textbf{0.567} \\
            \bottomrule
        \end{tabular}
    }
    \label{tab:manip_step1}
\vspace{-1em}
\end{table}

\vspace{-0.2em}
\noindent \textbf{Ablation on Reasoning Models.}
We evaluate multiple general-purpose models, both open-source and proprietary, for failure reasoning (Table~\ref{tab:manip_step1}). Among these, Gemini 2.5 Pro achieves the strongest semantic alignment with expert annotations, particularly under LLM-J scoring. While o4-mini and Cosmos-Reason1-7B achieve competitive ROUGE$_L$ and CS scores, their lower LLM-J agreement suggests weaker semantic correspondence despite surface-level textual similarity.

\noindent \textbf{Ablation on Downsampling Methods.}
We compare our semantic observation downsampling strategy against fixed-frame-rate sampling for failure explanation.
As shown in Table~\ref{tab:downsampling_methods}, our approach consistently outperforms fixed-rate baselines.
This indicates that embedding-based change-point selection better preserves semantically informative transitions within the VLM context window, leading to more accurate failure explanations.
\begin{table}[htbp]
    \centering
    \caption{Performance comparison with fixed frame rates sampling, highlighting benefits of the proposed semantic downsampling.  The best method is highlighted in \textbf{bold}.}
    \setlength{\tabcolsep}{6pt}
    \renewcommand{\arraystretch}{1.2}
    \resizebox{\columnwidth}{!}{%
        \begin{tabular}{c|c|c|c|c|c|c}
            \toprule
            \multirow{2}{*}{\textbf{Method}} 
            & \multicolumn{3}{c|}{\textbf{Simulation}} 
            & \multicolumn{3}{c}{\textbf{Real-World}} \\
            & \textbf{CS} $\uparrow$ 
            & \textbf{ROUGE$_{L}$} $\uparrow$ 
            & \textbf{LLM-J} $\uparrow$ 
            & \textbf{CS} $\uparrow$ 
            & \textbf{ROUGE$_{L}$} $\uparrow$ 
            & \textbf{LLM-J} $\uparrow$ \\
            \midrule
            1 fps    
            & 0.5927 & 0.2545 & 0.66 
            & 0.6141 & 0.2971 & 0.4166 \\
            0.5 fps  
            & 0.5885 & 0.2582 & 0.68 
            & 0.6362 & 0.3286 & 0.4332 \\
            0.25 fps 
            & 0.5917 & 0.2512 & 0.69 
            & \textbf{0.6581} & 0.3405 & 0.4333 \\
            \midrule
            \textbf{Ours} 
            & \textbf{0.6003} & \textbf{0.2589} & \textbf{0.76} 
            & 0.6280 & \textbf{0.3420} & \textbf{0.5667} \\
            \bottomrule
        \end{tabular}
        \label{tab:downsampling_methods}
    }
\end{table}

\vspace{0.2em}
\noindent \textbf{General-Purpose vs Fine-Tuned Models.}
We compare our approach against two manipulation-specific, instruction-fine-tuned VLMs: AHA-13B \cite{aha} and RoboFAC-7B \cite{lu2025robofac}. On the unseen RoboFail real-world dataset, our reasoning-based framework outperforms both models (Table~\ref{tab:robofail_real_world}).
While fine-tuned models can specialize to specific viewpoints and task distributions, we observe reduced generalization to unseen environments and hallucinated failure explanations. General-purpose reasoning models instead yield more robust cross-domain behavior without task-specific retraining, aligning with our goal of discovering deployment-scale failure taxonomies.

\begin{table}[htbp]
\vspace{0.5em}
    \centering
    \caption{Performance comparison of Gemini 2.5 Pro with fine-tuned models showing the benefits of general-purpose. Here, LLM-J is computed using the same model, prompt, and parameters as AHA-13B \cite{aha} for fair comparison.}
    \setlength{\tabcolsep}{4pt}
    \renewcommand{\arraystretch}{1.0}
    \resizebox{0.7\columnwidth}{!}{%
        \begin{tabular}{c|c|c|c}
            \toprule
            \textbf{Model} 
            & \textbf{CS} $\uparrow$ 
            & \textbf{ROUGE$_{L}$} $\uparrow$ 
            & \textbf{LLM-J} $\uparrow$ \\
            \midrule
            AHA-13B \cite{aha}              
            & 0.471 & 0.280 & 0.465 \\
            RoboFAC-7B \cite{lu2025robofac} 
            & 0.452 & 0.137 & 0.133 \\
            \midrule
            \textbf{Gemini 2.5 Pro} 
            & \textbf{0.628} & \textbf{0.342} & \textbf{0.550} \\
            \bottomrule
        \end{tabular}
    }
    \label{tab:robofail_real_world}
\vspace{-1.25em}
\end{table}

\begin{figure}[htbp]
  \vspace{-0.5em}
  \centering
  {\includegraphics[width=\columnwidth]{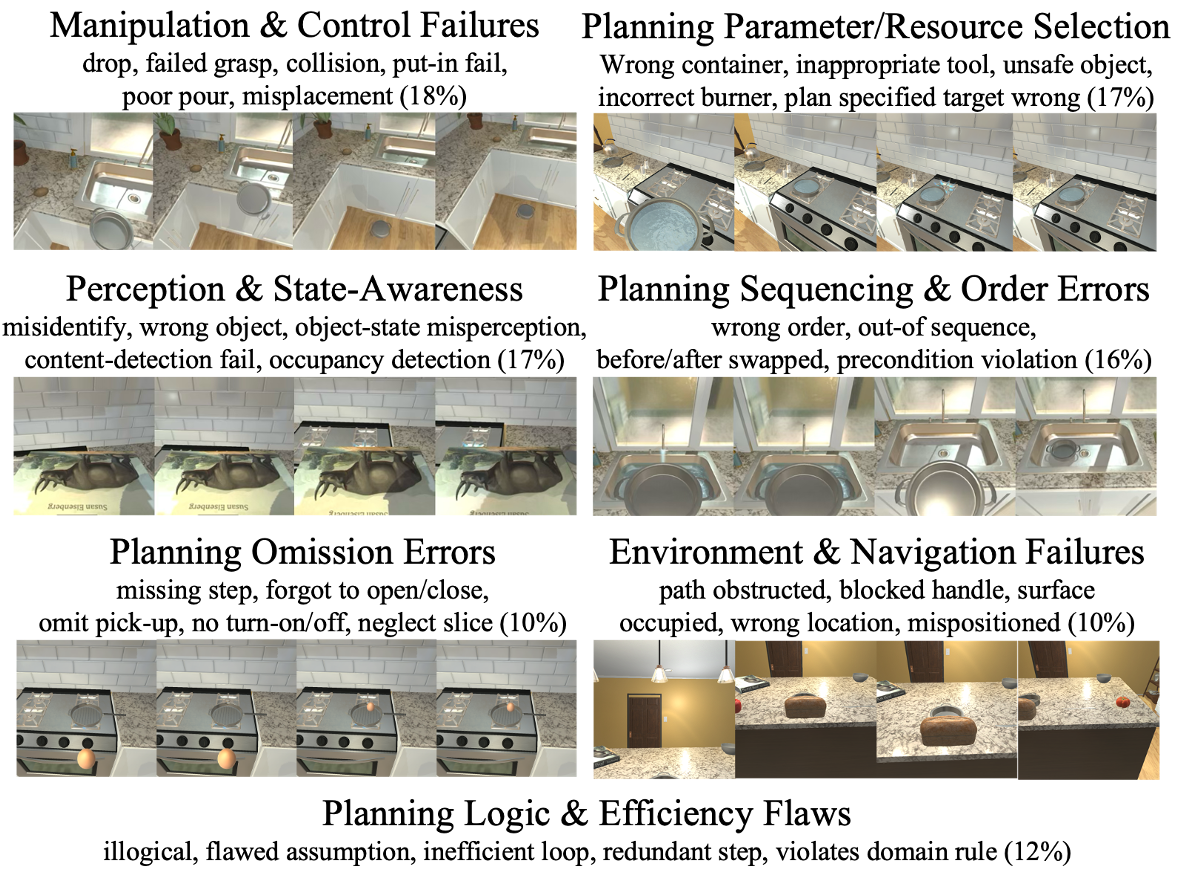}}
  \caption{Manipulation failure clusters with keywords and frequencies (under each cluster name) and examples.}
  \label{fig:cluster_visual_manip}
  \vspace{-1em}
\end{figure}

\paragraph{Validation of Taxonomy Recovery}
We next evaluate whether the obtained taxonomy recovers meaningful failure modes. From the failure explanations, our method produces semantically coherent clusters corresponding to distinct root causes. Fig.~\ref{fig:cluster_visual_manip} shows these clusters with representative keywords and frequencies.
Qualitatively, the taxonomy yields clear and interpretable groupings. For example, \texttt{Manipulation \& Control Failures} capture execution-level issues such as failed grasps and unintended drops; \texttt{Planning Parameter \& Resource Selection Errors} reflect incorrect or misspecified actions (e.g., wrong tool or container); and \texttt{Perception \& State Awareness Failures} isolate errors due to misidentification or incomplete scene understanding.
Overall, the clusters capture fine-grained distinctions across perception, planning, control, and environment-related failures, providing a rich, interpretable organization of the RoboFail failure dataset.

\begin{figure*}[t]
  \vspace{0.5em}
  \centering
  \centering
  \includegraphics[clip,width=0.99\textwidth]{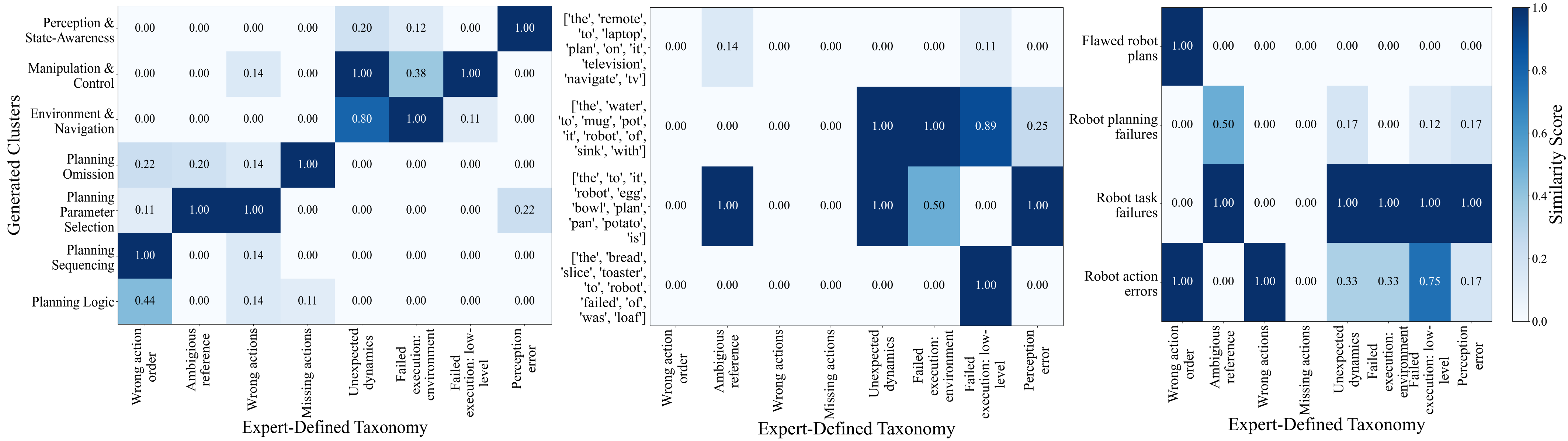}
  \caption{Heatmaps showing semantic similarity between the RoboFail expert-defined failure modes (columns) and the discovered clusters (rows) for our method (left), BERTopic (middle), and BERTopic-LLM (right). A sharper, diagonally dominant structure indicates stronger one-to-one alignment and better recovery of failure categories.}
  \label{fig:heatmaps}
  \vspace{-1em}
\end{figure*}

\vspace{0.2em}
\noindent \textbf{Baselines and Results.}
To evaluate the quality of the discovered failure clusters, we compare against BERTopic \cite{bertopic}, a state-of-the-art topic modeling method combining transformer embeddings with unsupervised clustering and keyword extraction, applied to the same failure explanations. We also include a stronger variant, BERTopic-LLM, which uses a language model to summarize each BERTopic cluster with representative keywords and descriptions.

\begin{table}[htbp]
\vspace{-0.5em}
  \centering
  \label{fig:bertopic_clusters}
  \small
  \begin{tabular}{@{} c >{\raggedright\arraybackslash}p{3cm} >{\raggedright\arraybackslash}p{4cm} @{}}
    \toprule
    \textbf{Size} & \textbf{BERTopic-LLM Clusters} & \textbf{BERTopic Keywords} \\
    \midrule
    (41) & Robot planning failures & the, water, to, mug, pot, it, robot, of, sink, with \\
    (38) & Robot task failures & the, to, it, robot, egg, bowl, plan, pan, potato, is \\
    (12) & Robot action errors & the, bread, slice, toaster, to, robot, failed, of, was, loaf \\
    (9)  & Flawed robot plans & the, remote, to, laptop, plan, on, it, television, navigate \\
    \bottomrule
  \end{tabular}
\end{table}

As evident from the list above, BERTopic groups frequent terms and fails to separate distinct underlying patterns. Although BERTopic-LLM improves interpretability, its clusters remain broad and overlapping, merging conceptually distinct categories (e.g., wrong action order, missing steps, and incorrect actions) into generic planning-related groups. It therefore does not reliably isolate the root-cause distinctions required for interpretable failure analysis.

For quantitative comparison, we prompt an LLM to score the semantic similarity of each pair of predicted and expert-defined failure mode from RoboFail's taxonomy. From the normalized similarity matrix $S \in [0,1]^{L \times N}$ (with $N{=}8$ expert failure modes and $L$ predicted clusters), we compute: (i) \textbf{Cluster Precision (CP)} $\left( \frac{1}{L}\sum_i \max_j S_{ij} \right)$, measuring whether each predicted cluster maps sharply to an expert failure mode; (ii) \textbf{Taxonomy Coverage (TC)} $\left( \frac{1}{N}\sum_j \max_i S_{ij} \right)$, measuring the completeness of the recovered taxonomy; and (iii) their harmonic mean, \textbf{Semantic Alignment Score (SAS)}, penalizing vague clusters and missed expert failure modes.

Fig.~\ref{fig:heatmaps} visualizes the similarity. Our method exhibits sharp, diagonally dominant structure, indicating near one-to-one alignment with the expert-defined taxonomy, whereas BERTopic variants produce diffuse, overlapping mappings. Table~\ref{tab:clustering_metrics} shows that consensus aggregation further improves CP, TC, and SAS relative to any single run and both BERTopic baselines, achieving TC\,=\,1.0 (covering all expert-defined failure modes). Together, these results demonstrate that our taxonomy is interpretable and expert-consistent while maintaining distinct categories.
\begin{table}[htbp]
\centering
\caption{Cluster quality metrics against the RoboFail expert taxonomy. Single Run reports the average over four independent runs. (CP: Cluster Precision, TC: Taxonomy Coverage, SAS: Semantic Alignment Score).}
\label{tab:clustering_metrics}
\setlength{\tabcolsep}{6pt}
\resizebox{0.7\columnwidth}{!}{
\begin{tabular}{lccc}
\toprule
\textbf{Method} & \textbf{CP} $\uparrow$ & \textbf{TC} $\uparrow$ & \textbf{SAS} $\uparrow$ \\
\midrule
BERTopic            & 0.785 & 0.625 & 0.696 \\
BERTopic-LLM            & 0.875 & 0.875 & 0.875 \\
\midrule
Ours (Single Run)   & 0.818 & 0.900 & 0.849 \\
\textbf{Ours (Aggregation)} & \textbf{0.920} & \textbf{1.000} & \textbf{0.958} \\
\bottomrule
\end{tabular}
}
\end{table}

\paragraph{Validation of Trajectory-to-Cluster Assignment}
Finally, we evaluate whether individual failure episodes can be reliably mapped to the obtained taxonomy, assigning each failure explanation to its most appropriate failure modes.
Our framework achieves a weighted F1 score of 85.53\%, significantly outperforming an embedding-similarity baseline that assigns each explanation to the cluster with the highest cosine similarity (F1 - 32.41\%).
These results demonstrate that the taxonomy is both discriminative and practically usable for consistent categorization.

\subsection{\textbf{Case Study 2: Real-World Car Crash Videos}}
\label{sec:driving_disc}

\begin{figure}[b]
  \vspace{-1em}
  \centering
  {\includegraphics[width=\columnwidth]{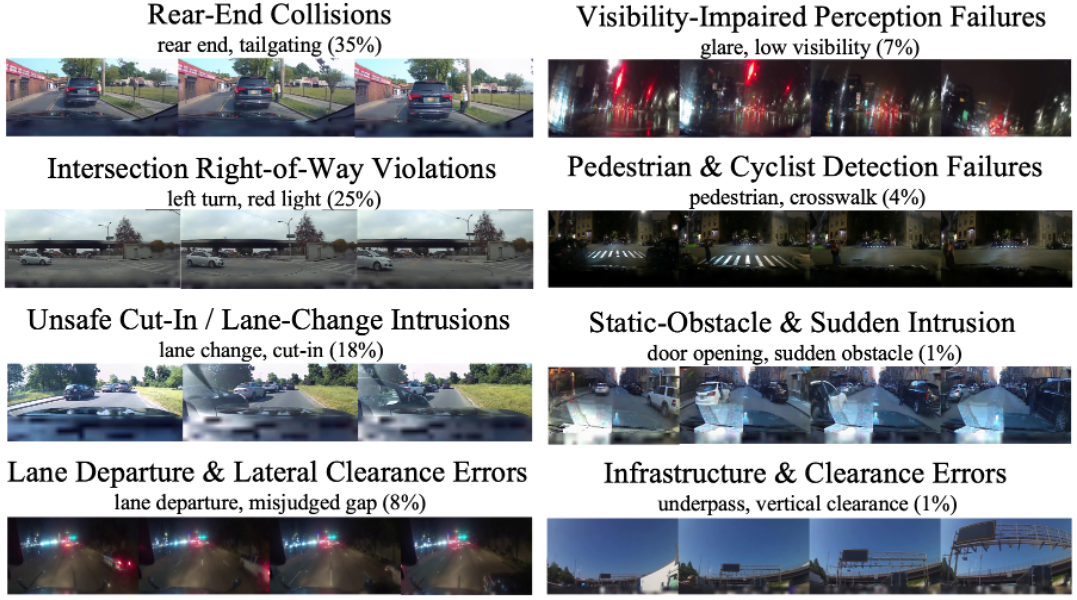}}
  \caption{Real-world car crash failure clusters with examples.}
  \label{fig:cluster_visual_driving}
  \vspace{-0.5em}
\end{figure}

We next evaluate our framework on the Nexar car crash dataset \cite{moura2025nexar}, containing 1,500 dashcam videos (approximately 40 seconds each at 1280×720 resolution and 30 fps) of collisions or near-miss events involving the ego vehicle. Although these recordings originate from human-driven vehicles, they provide a large-scale proxy for deployment-time driving failures in the absence of publicly available autonomous vehicle failure logs, to which the framework directly applies. Our method yields a semantically coherent taxonomy of recurring driving failure modes, illustrated in Fig.~\ref{fig:cluster_visual_driving}.

Qualitatively, the discovered clusters correspond to interpretable and recurring incident types, including \texttt{Rear End Collisions}, \texttt{Intersection Right of Way Violations}, and \texttt{Unsafe Cut In / Lane Change Intrusions}. The taxonomy also isolates rarer failures such as \texttt{Visibility Impaired Perception Failures} and \texttt{Static Obstacles}, demonstrating coverage across both frequent and long-tail scenarios.

Importantly, the taxonomy closely aligns with the U.S. DoT Volpe Center’s pre-crash typology \cite{crash_typology}, capturing major failure categories identified in traffic safety research. This alignment emerges without predefined labels or structured metadata, indicating that the framework can recover semantically grounded failure categories directly from raw deployment video logs.

\subsection{\textbf{Case Study 3: Vision-Based Indoor Robot Navigation}}
\label{sec:wayptnav_disc}
We apply our framework to a vision-based ground robot navigating previously unseen indoor office environments \cite{bansal2020combining}.
The robot employs a CNN-based policy with a model-based low-level controller, receiving RGB images, ego-velocities, and a goal position as inputs and outputting acceleration commands.
We record robot rollouts in the simulated Stanford office environment \cite{stanford_office} and extract front-view image sequences. Trajectories resulting in collisions constitute the failure dataset $D$.

\begin{figure}[htbp]
  \centering
  {\includegraphics[width=0.92\columnwidth]{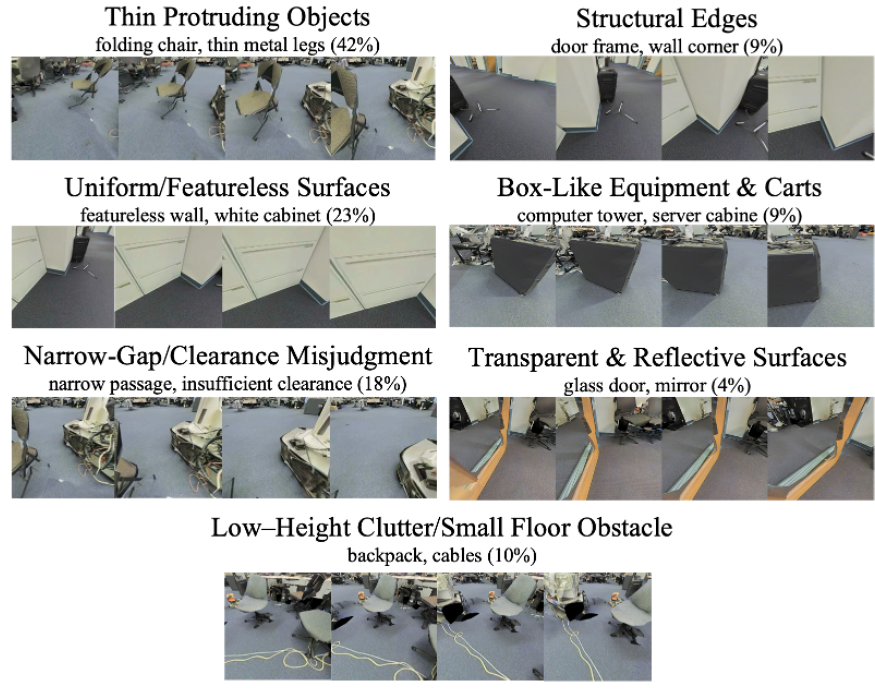}}
  \caption{Indoor navigation clusters with examples.}
  \label{fig:cluster_visual_wayptnav}
  \vspace{-0.5em}
\end{figure}

Applying our method to these collision trajectories yields a set of interpretable failure modes, shown in Fig.~\ref{fig:cluster_visual_wayptnav}. The clusters capture distinct perceptual and geometric failure causes, including \texttt{Thin Protruding Objects}, \texttt{Uniform/Featureless Surfaces}, and \texttt{Narrow Gap Misjudgments}.
Notably, several of them, such as protruding corners, glass doors, and featureless walls, align with failure types previously identified manually in \cite{chakraborty2023discovering}. This correspondence emerges without predefined labels, indicating that the framework can rediscover known causes while also organizing additional recurring collision patterns into a coherent taxonomy.

\section{Downstream Safety Improvement}
\label{sec:enhc}
To illustrate the utility of the discovered failure taxonomy, we present two proof-of-concept downstream integrations: runtime failure monitoring and targeted data collection. These experiments are not intended to introduce new failure detection or policy training pipelines, but to demonstrate how structured failure modes can be incorporated into existing frameworks to further improve robustness.

\subsection {Runtime Failure Monitoring}
Prior work in LLM-based anomaly detection monitors system safety by flagging unusual behavior in the scene~\cite{semanticanomaly}. We investigate a complementary hypothesis: providing system-specific failure modes can improve monitoring by enabling reasoning over known vulnerabilities.

We construct a runtime monitor that prompts a VLM with a history of observations and asks it to reason about potential safety violations. Crucially, we augment this prompt with the discovered failure modes, testing whether taxonomy-level context helps anticipate failures.

\vspace{0.2em}
\noindent \textbf{Baselines.}
To contextualize performance, we compare against representative anomaly-detection and supervised classification approaches. \textit{LLM-Based Anomaly Detection (LLM-AD)} \cite{semanticanomaly} provides human-written anomalous and nominal examples to an LLM and asks it to detect anomalies from scene descriptions. We also compare against supervised binary failure classifiers: \textit{VideoMAE-BC} \cite{wang2023videomae}, the top-performing model in the Nexar Crash Prediction Challenge \cite{moura2025nexar}, and \textit{ENet-BC} \cite{chakraborty2024enhancing}, trained on labeled indoor navigation collision data.
To isolate the contribution of taxonomy-level context, we further ablate our own method, removing failure cluster information from the monitoring prompt (\textit{NoContext}).
\begin{table}[t]
\vspace{0.5em}
\centering
\caption{Failure detection results for car crash videos and indoor robot navigation systems. We report F1 (\%age) scores on both In-D and OOD test samples, and the average lead detection time. The best method is highlighted in \textbf{bold}.}
\label{tab:fd_table}
\small
\setlength{\tabcolsep}{4pt}
\begin{tabularx}{\columnwidth}{Xccc}
\toprule
\textbf{Method} & \textbf{In-D. F1} $\uparrow$ & \textbf{OOD F1} $\uparrow$ & \textbf{Lead Time} $\uparrow$ \\
\midrule
\rowcolor{gray!15}
\multicolumn{4}{c}{Real-World Car Crash Videos} \\
\midrule
VideoMAE-BC \cite{wang2023videomae} & 65.3 & 25.2 & 506.6 ms \\
LLM-AD & 12.3 & 49.7 & 166.6 ms \\
NoContext & 54.1 & 69.6 & 473.3 ms \\
\midrule
\textbf{Ours} & \textbf{71.4} & \textbf{77.9} & \textbf{610 ms} \\
\midrule
\rowcolor{gray!15}
\multicolumn{4}{c}{Vision-Based Indoor Robot Navigation} \\
\midrule
ENet-BC \cite{chakraborty2024enhancing} & \textbf{78.8} & 22.4 & 1.01 sec \\
LLM-AD & 40.0 & 27.2 & \textbf{1.38 sec} \\
NoContext & 67.4 & 40.5 & 0.76 sec \\
\midrule
\textbf{Ours} & 77.2 & \textbf{50.0} & 1.21 sec \\
\bottomrule
\end{tabularx}
\vspace{-1.5em}
\end{table}

\vspace{0.2em}
\noindent \textbf{Results.}
We evaluate all methods on 200 in-distribution (In-D) driving videos and 200 out-of-distribution (OOD) videos from a separate crash dataset. As shown in Table~\ref{tab:fd_table}, augmenting the monitor with the discovered failure clusters improves the F1 score, suggesting that taxonomy-level context enables more precise differentiation between nominal but rare behavior and structurally unsafe scenarios.
On OOD samples, our monitor maintains stronger generalization than supervised classifiers trained on specific datasets. It also detects failures earlier by correlating unfolding observations with known failure modes.
 
For indoor navigation, on an In-D test set of 326 trajectories, the cluster-guided monitor again outperforms LLM-AD in F1 score and matches the environment-specific ENet-BC classifier (Table~\ref{tab:fd_table}). On 300 OOD trajectories from a different building, all methods degrade, but the cluster-augmented approach maintains the highest F1 score while ENet-BC fails to generalize beyond its training distribution.

\begin{figure}[h]
  \centering
  {\includegraphics[width=\columnwidth]{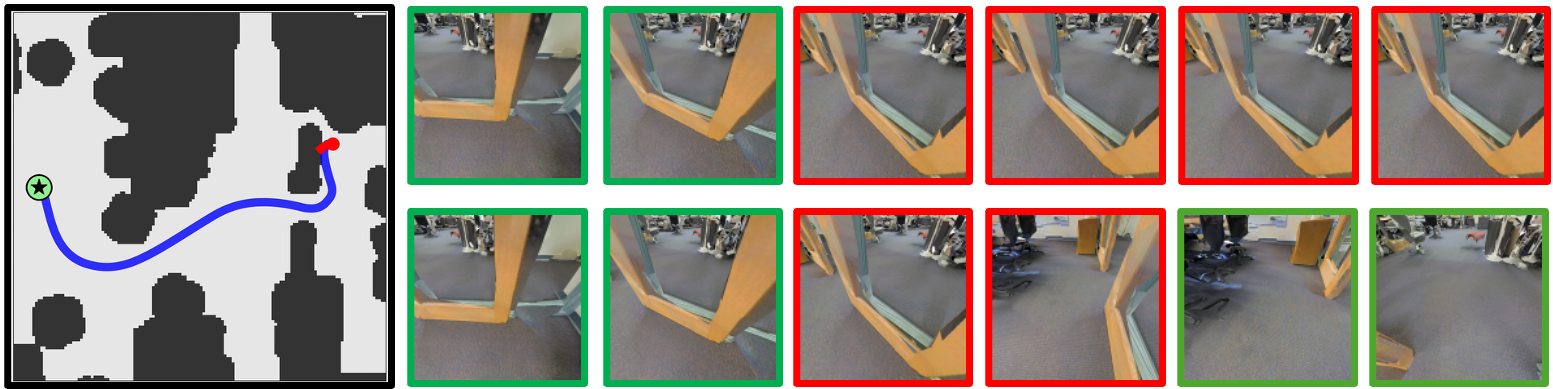}}
  \caption{Robot failing under nominal policy (red) due to misidentifying glass door as traversible, but succeeding under safeguard policy (blue). Red and green borders around FPV images indicate unsafe and safe predictions, respectively.}
  \label{fig:fallback}
  \vspace{-1em}
\end{figure}

\vspace{0.2em}
\noindent \textbf{Closed-Loop Safeguard Integration.}
To demonstrate closed-loop intervention, we integrate the runtime monitor with a reactive safeguard controller. Fig.~\ref{fig:fallback} shows a scenario where the nominal policy misidentifies a glass door as traversable. The taxonomy-guided monitor recognizes structural similarity to a known failure mode and triggers a safeguard policy (a geometric planner), preventing collision.
This illustrates how failure taxonomies can serve as actionable safety signals within existing control architectures.

\subsection{Targeted Data Collection and Policy Refinement}
As a second illustrative integration, we examine how the obtained failure taxonomy can guide targeted data collection for policy refinement, evaluating whether discovered failure patterns help identify the regions of the environment that would most benefit from additional supervision.
Prior work has shown that specification-guided or failure-aware data collection can improve robustness \cite{chakraborty2024enhancing, gupta2024detecting, dreossi2018counterexample, shah2023specification}. Our framework is complementary: it leverages the discovered semantically grounded failure modes to direct data collection without requiring any predefined specifications.

For the indoor navigation system, we use the discovered failure modes to identify high-risk regions of the environment and collect additional demonstrations there, for example featuring featureless walls and thin objects. The policy is then fine-tuned on the dataset containing 40K additional samples.
Consequently, failure rates in sampled trajectories reduce from 46\% to 18\%, demonstrating improved robustness in previously failure-prone regimes.
In contrast, fine-tuning with an equal number of uniformly collected samples reduces the failure rate only to 34\%, indicating that failure-guided data collection yields more efficient safety improvements.
This experiment illustrates how discovered failure taxonomies can support iterative system improvement by informing where additional supervision is most impactful.

\section{Conclusion and Future Work}
\label{sec:conclusion}

We present a framework for unsupervised discovery of

\noindent failure modes from multimodal deployment logs. By transforming raw failure trajectories into structured explanations and organizing them into recurring, semantically coherent categories, the proposed approach discovers interpretable failure taxonomies without requiring predefined labels or task-specific engineering. 
These taxonomies provide actionable structure for downstream safety workflows. Through proof-of-concept integrations, we show that failure taxonomy context can guide targeted data collection, inform policy refinement in high-risk regimes, and improve runtime failure monitoring.

Despite these advances, several limitations remain. 
First, there is no single canonical failure taxonomy, and different clustering or reasoning strategies may surface complementary perspectives on the same dataset. Second, while vision-language models enable scalable semantic interpretation, they may produce plausible but incorrect explanations. Future work could incorporate causal validation, simulation-based verification, or formal safety analysis (e.g., STPA or FRAM) to further ground failure modes. 
Finally, the pipeline’s modular design inherently supports scaling to larger datasets. Future work will investigate hierarchical and iterative clustering architectures with persistent context memory, enabling the framework to efficiently accommodate extensive, temporally extended deployment logs.

\renewcommand*{\bibfont}{\footnotesize}  
\printbibliography

\end{document}